\documentclass{article}
\usepackage{spconf,amsmath,graphicx}
\usepackage{tabularx}
\usepackage{booktabs}
\usepackage{amssymb}
\usepackage{array}
\usepackage{arydshln}

\title{Prompting Large Language Models for Zero-Shot Domain Adaptation \\ in Speech Recognition}

\twoauthors
 {Yuang Li\sthanks{Work done during an internship at Microsoft.}}
	{University of Cambridge}
 {Yu Wu, Jinyu Li, Shujie Liu}
 {Microsoft}

\begin{document}
%
\maketitle
\begin{abstract}
The integration of Language Models (LMs) has proven to be an effective way to address domain shifts in speech recognition. However, these approaches usually require a significant amount of target domain text data for the training of LMs. 
Different from these methods, in this work, with only a domain-specific text prompt, we propose two zero-shot ASR domain adaptation methods using LLaMA, a 7-billion-parameter large language model (LLM).  LLM is used  in two ways: 1) second-pass rescoring: reranking N-best hypotheses of a given ASR system with LLaMA; 2) deep LLM-fusion: incorporating LLM into the decoder of an encoder-decoder based ASR system. Experiments show that, with only one domain prompt, both methods can effectively reduce word error rates (WER) on out-of-domain TedLium-2 and SPGISpeech datasets. Especially, the deep LLM-fusion has the advantage of better recall of entity and out-of-vocabulary words.
\end{abstract}


\section{Introduction}

\begin{table}[ht]
    \centering
    \small
    \begin{tabular}{c|p{6cm}} 
        \hline
        \textbf{prompt 1} & The following text is the transcription of company earnings calls.\\
        \hdashline
        \textbf{prompt 2} & toy tiles that talk to each other\\
        \hdashline
        \textbf{prompt 3} & dreams from endangered culture\\
        \hline
        \multicolumn{2}{c}{}\\
        \hline
        reference & \textbf{uptick} we're seeing in the \textbf{containerboard} market\\
        \hdashline
        no prompt & \textbf{optic} we're seeing in the \textbf{container board} market\\
        \hdashline
        \textbf{+ prompt 1} & \textbf{uptick} we're seeing in the \textbf{containerboard} market\\
        \hline
        reference & well \textbf{opportunistic} share \textbf{repurchases} are clearly at the top of that list\\
        \hdashline
        no prompt & well \textbf{oppertunistic} share \textbf{re purchases} are clearly at the top of that list\\
        \hdashline
        \textbf{+ prompt 1} & well \textbf{opportunistic} share \textbf{repurchases} are clearly at the top of that list\\
        \hline
        reference & now this is an \textbf{interactive} \textbf{cartoon} application\\
        \hdashline
        no prompt & now this is an \textbf{iniractive} a \textbf{cartune} application\\
        \hdashline
        \textbf{+ prompt 2} & now this is an \textbf{interactive} a \textbf{cartoon} application\\
        \hline
        reference & the myths of the \textbf{inuit} elders still resonate with meaning or that in the \textbf{himalaya} the\\
        \hdashline
        no prompt & the myths of the \textbf{innuit} elders still resonate with meaning or that in the \textbf{himalaia} the\\
        \hdashline
        \textbf{+ prompt 3} & the myths of the \textbf{inuit} elders still resonate with meaning or that in the \textbf{himalaya} the\\
        \hline
    \end{tabular}
    \caption{Examples of deep LLM-fusion with/without prompt.}
    \label{table:example}
\end{table}

End-to-end (E2E) automatic speech recognition (ASR) systems~\cite{graves2006connectionist, graves2012sequence, chorowski2015attention, radford2022robust, E2EOverview} have demonstrated superior performance over traditional pipeline approaches, with increases in model size and the scale of the supervised dataset. However, E2E ASR still suffers from domain mismatch and limited utilization of text corpora. To address these issues, external Language Models (LMs) are often incorporated. Among them, two techniques without altering the ASR model architecture are second-pass rescoring~\cite{zheng2021adapting} and shallow fusion\cite{kannan2018analysis}. Alternatively, LMs can be integrated into the ASR model decoder as internal LMs, such as deep fusion\cite{gulcehre2015using} and cold fusion~\cite{sriram2018cold}, merging the hidden states of the LM with the ASR model. Factorized neural transducer model ~\cite{chen2022factorized, zhao2023fast} is another promising architecture that predicts the blank token and vocabulary tokens separately so that the vocabulary predictor fully functions as an LM.

Leveraging LMs can make ASR domain adaptation more accessible since it is easier to collect target domain text than audio-text pairs. In this case, an LM used in ASR adaptation can be developed through fine-tuning~\cite{meng2021internal, meng2021internal2, meng2021internal3, tsunoo2022residual, deng2023adaptable} or prompt tuning~\cite{dingliwal2021domain}. Although these approaches have shown promising results, it is worth noting that commonly used LMs, such as GPT-2~\cite{radford2019language} and Transformer-XL~\cite{dai2019transformer}, have relatively small scales and lack in-context learning capability. In contrast, the latest Large Language Models (LLMs), such as GPT-4~\cite{gpt2023gpt} and LLaMA~\cite{touvron2023LLaMA}, offer significantly larger capacities. These LLMs can be adapted to downstream tasks without training by adding a textual description of the task as a prompt~\cite{brown2020language}. Therefore, LLMs offer advantages for domain adaptation, including:

\begin{itemize}
    \item \textbf{Zero-shot adaptation}: The traditional approach of acquiring additional audio/text data for domain adaptation is not only time-consuming but costly, and sometimes, it can be impossible to obtain domain-specific data. By prompting, we can adapt LLM to new domains and use it in ASR without re-training.
    \item \textbf{Flexible prompt design}: The LLM possesses an exceptional capability to extract information from texts of varying forms and lengths, making prompt design effortless. Depending on the situation, various types of prompts can be utilized effectively. For instance, a concise summary of a talk can be provided to enhance recognition accuracy, or alternatively, only a title or important entity words can be used as prompts.
    \item \textbf{Leverage rapid-growth LLM community}: In recent years, the capabilities of LLMs have rapidly evolved, with tremendous breakthroughs observed in both open-source models and business APIs. We are confident that the proposed framework for addressing ASR domain shift will continue to improve with the progress of LLMs.
\end{itemize}


In this work, we utilized LLaMA-7B and handcrafted textual prompts to adapt ASR models trained on LibriSpeech~\cite{panayotov2015librispeech} to TedLium-2~\cite{rousseau2012ted} and SPGISpeech~\cite{o2021spgispeech} datasets. The prompts were derived from the video description of TED talks and the dataset description of SPGISpeech. Initially, we employed LLaMA to rerank the 16-best hypotheses of the HuBERT-CTC model~\cite{hsu2021HuBERT}. When prompts were incorporated, we observed relative word-error-rate (WER) reductions of 6.6\% and 3.3\% on the TedLium-2 and SPGISpeech datasets, respectively. The effectiveness of the second-pass reranking is constrained by the quality of N-best hypotheses. Therefore, in further experiments, LLaMA was directly incorporated into an encoder-decoder framework, which was modified from Flamingo~\cite{alayrac2022flamingo} and referred to as deep LLM-fusion in this paper. In this setup, acoustic features from the HuBERT model were fed into the frozen LLaMA model using the Gated cross-attention mechanism. By using domain prompts, we achieved 2.6\% and 7.7\% relative WER reductions on the TedLium-2 and SPGISpeech datasets, respectively. Notably, domain prompts enabled more accurate recognition of semantically important entities, as demonstrated in Table~\ref{table:example}. For example, with a prompt indicating the recording is company earning calls, the word "uptick", which describes a slight upward trend, can be recognized accurately instead of being mistakenly recognized as "optic."

\section{Related Work}

\subsection{Domain Adaptation through Internal LM}

The E2E ASR model learns an internal LM from the training data, but it is biased toward the source domain. To address this bias when transferring to a new domain, density ratio ~\cite{mcdermott2019density},  hybrid autoregressive transducer \cite{variani2020hybrid}, internal LM estimation (ILME)~\cite{meng2021internal, meng2021internal2} methods predict the internal LM score and subtracts it from the combined scores of the E2E model and the external LM. However, these methods introduce additional complexity to the decoding process, and the estimation of the internal LM can be inaccurate. Therefore alternative approaches were proposed, which involve fine-tuning the internal LM~\cite{chen2022factorized, meng2021internal3} or replacing it with a target-domain LM~\cite{deng2023adaptable}. This allows the ASR model to be directly adapted to a new domain without an external LM. In this work, the deep LLM-fusion method can be considered as explicitly using LLaMA as an internal LM.

\subsection{Domain Adaptation through Prompting}

Prompting, which involves adding prefixes to the text input, is a popular method for LM adaptation~\cite{li2021prefix}. It can be categorized into discrete and continuous prompts. In~\cite{shenoy2021adapting, shenoy2021asr}, recurrent and Transformer-XL-based LMs with discrete prompts were used to rescore the N-best hypotheses of the ASR model. The prompt comprises historical transcriptions and dialogue acts, obtained from the output of the natural language understanding model. This approach biases the LM towards the domain indicated by the prompt context. To eliminate the need for manual prompt design, prompt tuning~\cite{dingliwal2021domain} employs learnable weights as prompts. It requires training in the target domain, but the number of updated parameters is substantially reduced compared to fine-tuning the entire LM. In our study, we designed straightforward descriptive prompts for the TedLium-2 and SPGISpeech datasets, and employed a significantly larger LM, surpassing the scale of previous approaches.

\subsection{From LLM to Multi-Modal LLM}

The adaptation of LLM to multi-modal tasks has been a recent research hotspot, with a particular focus on visual understanding tasks. One approach was proposed in MiniGPT-4~\cite{zhu2023minigpt}, which directly feeds visual features into the LLM after a single projection layer for alignment. Another approach, LLaMA-Adapter~\cite{zhang2023LLaMA} utilizes fixed-length trainable vectors as layer-wise prompts which can incorporate visual information during the instructive fine-tuning. Both MiniGPT-4 and LLaMA-Adapter are decoder-only models. On the other hand, Flamingo~\cite{alayrac2022flamingo} employs an encoder-decoder framework where visual representations are fused into the LLM through cross-attention. In all of these methods, the LLM remains frozen, and only a small number of parameters are introduced. To stabilize training, visual features are gradually incorporated using techniques such as zero-init attention in LLaMA-Adapter and zero-init gating in Flamingo. In this work, deep LLM-fusion can be viewed as adapting LLM to speech modality. Considering the need to handle long and variable-length input, our architecture is similar to Flamingo.

\begin{figure*}[ht]
\centering
  \includegraphics[width=0.9\linewidth]{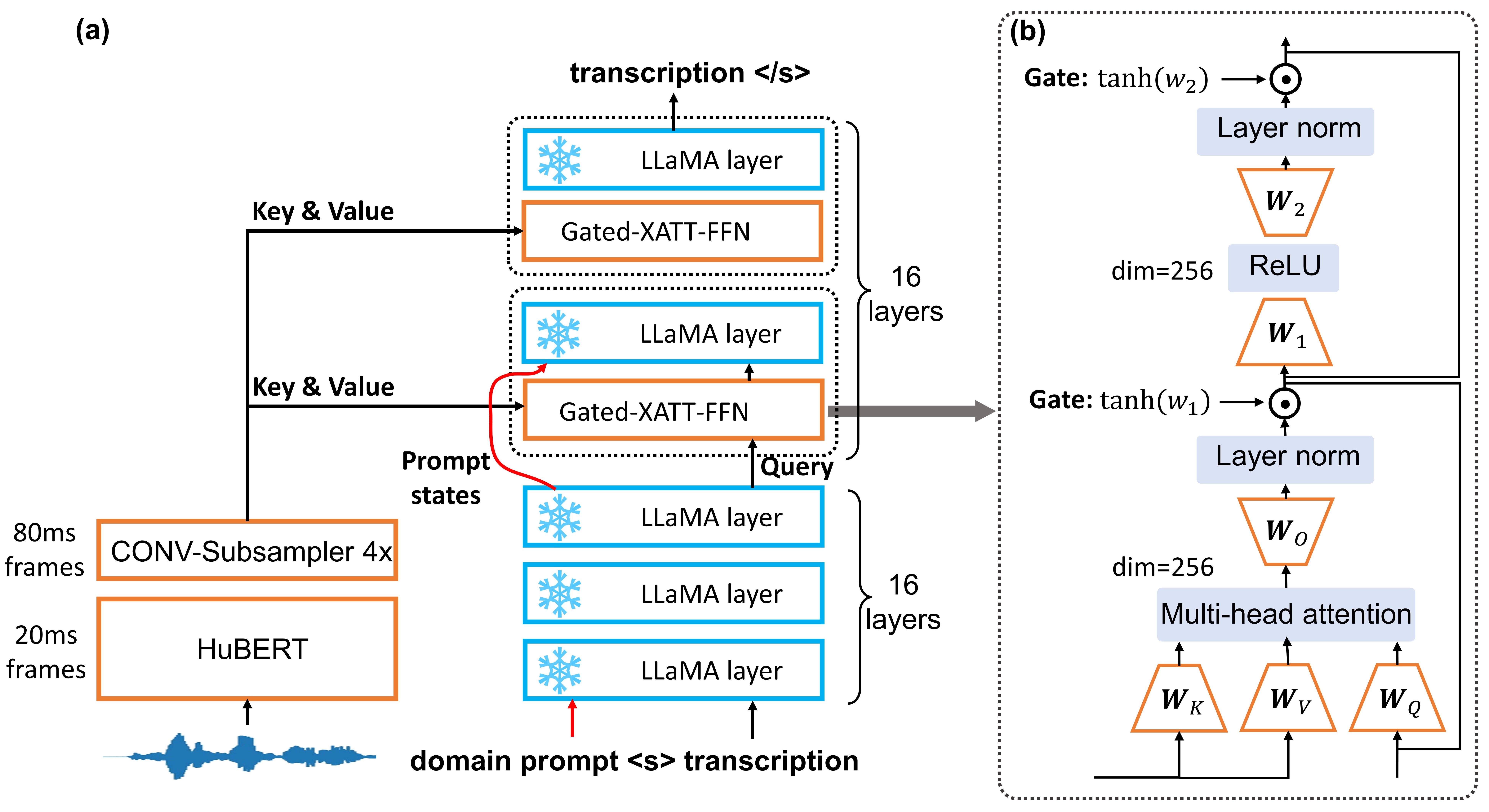}
  \caption{(a) Deep fusion of LLaMA into an attention encoder-decoder-based ASR model with HuBERT as the encoder. (b) Illustration of Gated-XATT-FFN module.}
  \label{fig:fusion}
\end{figure*}

\section{Method}

\subsection{Second-pass Reranking}

\begin{figure}[h]
\centering
  \includegraphics[width=\linewidth]{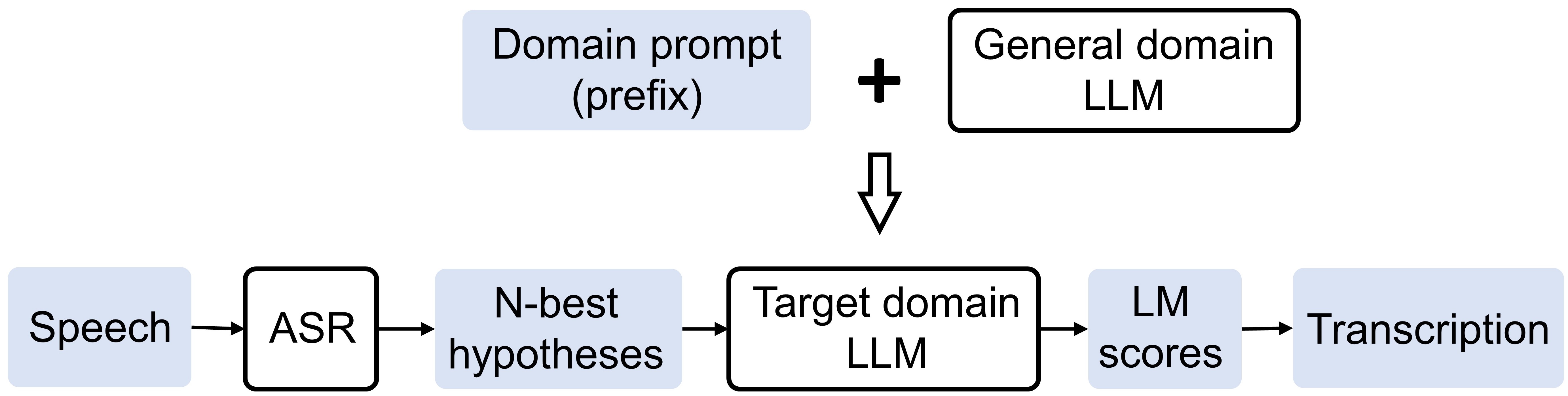}
  \caption{Flowchart for the second-pass reranking.}
  \label{fig:rerank}
\end{figure}

Figure~\ref{fig:rerank} illustrates the flowchart of the second-pass reranking method for domain adaptation. An ASR model first generates N-best hypotheses. Then, the LLaMA model with domain prompt, computes the LM score (Equation~\ref{eq:1}) as the sum of log probabilities for each hypothesis. In essence, a domain-specific textual prompt is added to the beginning of each hypothesis before rescoring. Finally, the hypothesis with the highest LM score is chosen. Note that we did not combine LM score with E2E ASR score due to limited benefits. Furthermore, shallow fusion was not considered as the vocabulary mismatch between the ASR model and LLM complicates the decoding process. 

\begin{align}
\label{eq:1}
\text{LM score} = \sum_{i=1}^{N} \log P(w_i | \mathbf{w}_{<i}, \mathbf{w}_{p})
\end{align}

\noindent where the probability of current wordpiece $w_i$ is conditioned on previous wordpieces $\mathbf{w}_{<i}$ and wordpieces in the prompt $\mathbf{w}_p$.

\subsection{Deep LLM-Fusion}

\subsubsection{Model Architecture}

Figure~\ref{fig:fusion} (a) illustrates the integration of LLaMA into a sequence-to-sequence model using gated cross attention and feed-forward network (Gated-XATT-FFN) modules~\cite{alayrac2022flamingo}. Initially, the raw waveform input is processed by the HuBERT-Large model, which is followed by a convolutional subsampling module that reduces the acoustic feature from 20ms per frame to 80ms per frame. These features serve as keys and values in the Gated-XATT-FFN module, while the hidden states from LLaMA act as queries. Finally, transcriptions are generated using the causal LLaMA decoder. During training, the LLaMA layers remain fixed whereas all other modules are updated. This architecture was adapted from Flamingo~\cite{alayrac2022flamingo} and the key differences are:

\begin{itemize}
\setlength{\parskip}{0pt}
    \item The speech encoder replaces the visual encoder and it was finetuned instead of frozen. We also tried to unfreeze the last layer of LLaMA, contributing to further performance improvements.
    \item To enable the LLaMA decoder to distinguish the prompt and the transcription, a special start-of-sentence token was added between them. Additionally, the hidden states corresponding to the prompt skip the Gated-XATT-FFN module so that the first hidden state corresponding to the special token directly queries the beginning of acoustic features.
    \item The computational requirement was reduced by decreasing the temporal dimension of acoustic features and the hidden dimension of Gated-XATT-FFN modules. In addition, Gated-XATT-FFN modules were only included in the top layers of LLaMA, eliminating the need to store gradients for shallow layers.
\end{itemize}

\subsubsection{Low-Dimensional Gated Cross Attention}

The Gated-XATT-FFN module in Figure~\ref{fig:fusion} (b) is a key component for integrating the speech modality into LLaMA. It employs a Tanh gating mechanism (Equation~\ref{eq:2}~\ref{eq:3}), which determines the extent of influence that speech features have on the final transcriptions. The module consists of two trainable gating parameters, $w_1$ and $w_2$, responsible for controlling the cross-attention and FFN, respectively. To ensure a stable training process, these parameters are initialized as zero, so speech features are gradually incorporated.

 \begin{align}
 \label{eq:2}
\mathbf{Y} &= tanh(w_1) \odot MHA(\mathbf{K}, \mathbf{V}, \mathbf{Q}) + \mathbf{Q} \\
\label{eq:3}
\mathbf{\hat{Y}} &= tanh(w_2) \odot FFN(\mathbf{Y}) + \mathbf{Y}
\end{align}

Traditional cross-attention modules usually maintain a consistent dimension during the computation. Since LLaMA-7B has a large hidden dimension of 4096, a single linear projection matrix will have 16 million parameters. To decrease the scale of the Gated-XATT-FFN module, we implemented a bottleneck structure by reducing the hidden dimension inside the multi-head attention and FFN to 256.

\subsubsection{Long-form Training}

During inference, the LLaMA decoder utilizes a prompt to guide the autoregressive generation of transcriptions in the target domain. To enable this, it is crucial to have contextual prefixes before each transcription during training. These prefixes need to be diverse to enhance generalizability. However, manually constructing prompts for individual utterances is not feasible, and using the same prompt for multiple utterances is not ideal. Therefore, we adopted long-form training by using a single ground-truth transcription of the previous utterance as the prompt. By doing so, we encouraged the model to effectively incorporate the historical text, which encompasses local context or topic-related information.

\section{Experimental Setup}

\subsection{Model Configurations}

In our experiments, we utilized the LLaMA model, which has 7 billion parameters, 32 layers, and a hidden dimension of 4096. For the second-pass reranking, we employed the open-source HuBERT-Large model\footnote{https://huggingface.co/facebook/hubert-large-ls960-ft} that was pre-trained on the LibriLight~\cite{kahn2020libri} dataset and finetuned on the LibriSpeech~\cite{panayotov2015librispeech} dataset using a character-level vocabulary and CTC loss. During the decoding process, we used a beam size of 16 for all test sets. After rescoring with LLaMA, we selected the hypothesis with the highest LM score. For deep LLM-fusion, we utilized the HuBERT-Large model\footnote{https://huggingface.co/facebook/hubert-large-ll60k}, pre-trained on the LibriLight dataset, as a speech encoder. It takes 16kHz raw waveform as input and generates hidden states at a frame rate of 50Hz. To downsample the hidden states, we applied two 1D convolutional layers with a kernel size of three and a stride of two, resulting in a frame rate of 12.5Hz. These downsampled states were employed as keys and values in 16 Gated-XATT-FFN modules, which were inserted in the top 16 layers of LLaMA. In each Gated-XATT-FFN module, the multi-head cross attention has a dimension of 256 and four heads.

The deep LLM-fusion model was trained on the LibriSpeech dataset with Specaug~\cite{park2019specaugment} that masks HuBERT states and a three-phase training schedule. In the first phase, only the CONV-Subsampler and Gated-XATT-FFN modules were trained from scratch for 200,000 steps with AdamW~\cite{loshchilovdecoupled} optimizer, a batch size of 64, and cross-entropy loss. An inverse square root schedule was used with a peak learning rate (LR) of 1e-4. In this phase, the number of trainable parameters was 82 million. In the second phase, the HuBERT model was unfrozen and finetuned for another 100,000 steps with an LR of 3e-5. The number of learnable parameters increased to 475 million. It's worth noting that in the first two phases, the model was trained on the utterance-wise ASR task. In the third phase, history transcriptions were incorporated as prompts with a probability of 80\%, and the model was fine-tuned for an additional 100,000 steps with an LR of 1e-5. In this phase, it is optional to unfreeze the last LLaMA layer to enhance performance, which introduces another 460 million parameters. The training process utilized eight NVIDIA Tesla V100 32GB GPUs, with an actual batch size of 1 on each GPU. A simulated batch size of 64 was achieved by accumulating gradients over eight steps. During decoding, beam search was employed with a beam size of 16.


\subsection{Evaluation}

\subsubsection{Datasets}


ASR models trained on the LibriSpeech dataset~\cite{panayotov2015librispeech} were evaluated on the TedLium-2~\cite{rousseau2012ted} and SPGISpeech datasets~\cite{o2021spgispeech}. The LibriSpeech training set contains 960 hours of speech data. For the SPGISpeech eval set, we selected a subset of 15 hours of speech. Each sample in this subset has at least two out-of-vocabulary (OOV) words compared to the LibriSpeech vocabulary. The prompt used for the SPGISpeech eval set is: "The following text is the transcription of company earning calls." For the TedLium-2 dataset, we combined the dev and eval sets, resulting in recordings of 20 TED Talk videos which have 4 hours of speech. We collected the titles and the descriptions of each video as domain-specific prompts (Table~\ref{table:prompt}). All prompts are normalized by removing punctuations and changing words to lowercase, so the format is the same as history transcription used in long-form training. Additionally, long-form ASR performance using one history transcription will be provided for LibriSpeech and TedLium-2 datasets since their utterances were derived from long-from recordings.

\begin{table}[ht]
    \centering
    \begin{tabular}{p{2cm}|p{4cm}}
        \hline
         & \textbf{Prompt}\\
        \hline
        TedLium-2\newline (title) & Dreams from endangered cultures\\
        \hline
        TedLium-2\newline (description) & With stunning photos and stories, National Geographic Explorer Wade Davis celebrates the extraordinary diversity of the world's indigenous cultures, which are disappearing from the planet at an alarming rate.\\
        \hline
        SPGISpeech (description) & The following text is the transcription of company earnings calls.\\
        \hline
    \end{tabular}
    \caption{Examples of prompts.}
    \label{table:prompt}
\end{table}


\subsubsection{Metrics}

For LibriSpeech test sets, standard word-error-rate (WER) was used. For domain adaptation, we also considered the recall of entity words and out-of-vocabulary (OOV) words, which carry more semantic meaning. To achieve this, we employed a name entity recognition model to extract entity words~\footnote{https://huggingface.co/dslim/bert-large-NER} from both the ground-truth labels and the hypotheses. The recall was then calculated as the ratio of the recovered entity words to the total number of entity words. Similarly, OOV recall was determined by extracting OOV words using the LibriSpeech vocabulary.
\section{Results}

\subsection{Results for Second-pass Reranking}





Second-pass reranking generally benefits ASR adaptation, as demonstrated in Table~\ref{table:rarank2}. On the SPGISpeech dataset, reranking contributes to a 1.98\% absolute reduction in WER. Furthermore, by prepending the dataset description to each hypothesis before reranking, we achieved an additional 0.39\% absolute improvement in WER. We also observed significant enhancements in entity and OOV recall, with 6\% and 10\% absolute increases compared to the CTC baseline respectively. On the TedLium-2 dataset, reranking without a prompt did not noticeably improve the WER. When using a short title prompt, we observed a modest 0.23\% absolute improvement in WER. Longer prompts and incorporating local history context resulted in more substantial performance enhancements, with up to a 0.58\% absolute reduction in WER. While entity recalls only saw slight improvements, OOV recall was improved more substantially with an absolute increase of 7\%. In summary, the benefits of second-pass reranking depend on how well LLaMA aligns with the target domain, which can be enhanced through prompting. Additionally, the upper-bound performance is determined by the quality of the N-best hypothesis, which explains the limited improvements of the entity recall on the TedLium-2 dataset which has an upper bound of 0.56.

\begin{table}[ht]
\centering
\begin{tabular}{c |c | c | c  c} 
 \hline
 & & & \multicolumn{2}{c}{\textbf{Recall}}\\
\textbf{Reranking} & \textbf{Prompt} & \textbf{WER} & \textbf{Entity} & \textbf{OOV}\\
\hline
\multicolumn{5}{c}{\textit{SPGISpeech Dataset}} \\
\hline
\texttimes & \texttimes & 13.66 & 0.50 & 0.48 \\
\checkmark & \texttimes & 11.68 & 0.55 & 0.56 \\
\checkmark & description & \textbf{11.29} & \textbf{0.56} & \textbf{0.58} \\ 
\hline
\multicolumn{5}{c}{\textit{TedLium-2 Dataset}} \\ 
\hline
\texttimes & \texttimes & 9.22 & 0.50 & 0.38 \\
\checkmark & \texttimes & 9.25 & 0.49 & 0.44 \\
\checkmark & title & 8.99 & 0.50 & 0.45 \\
\checkmark & description & \textbf{8.64} & 0.51 & 0.45 \\
\checkmark & history-gt & \textbf{8.64} & 0.51 & 0.45 \\
\checkmark & history-hyp & 8.65 & 0.51 & 0.45 \\
 \hline
\end{tabular}
\caption{Domain adaptation performance of HuBert-CTC with/without LLaMA reranking.}
\label{table:rarank2}
\end{table}

\subsection{Results for Deep LLM-Fusion}

\begin{figure}[ht]
\centering
  \includegraphics[width=0.8\linewidth]{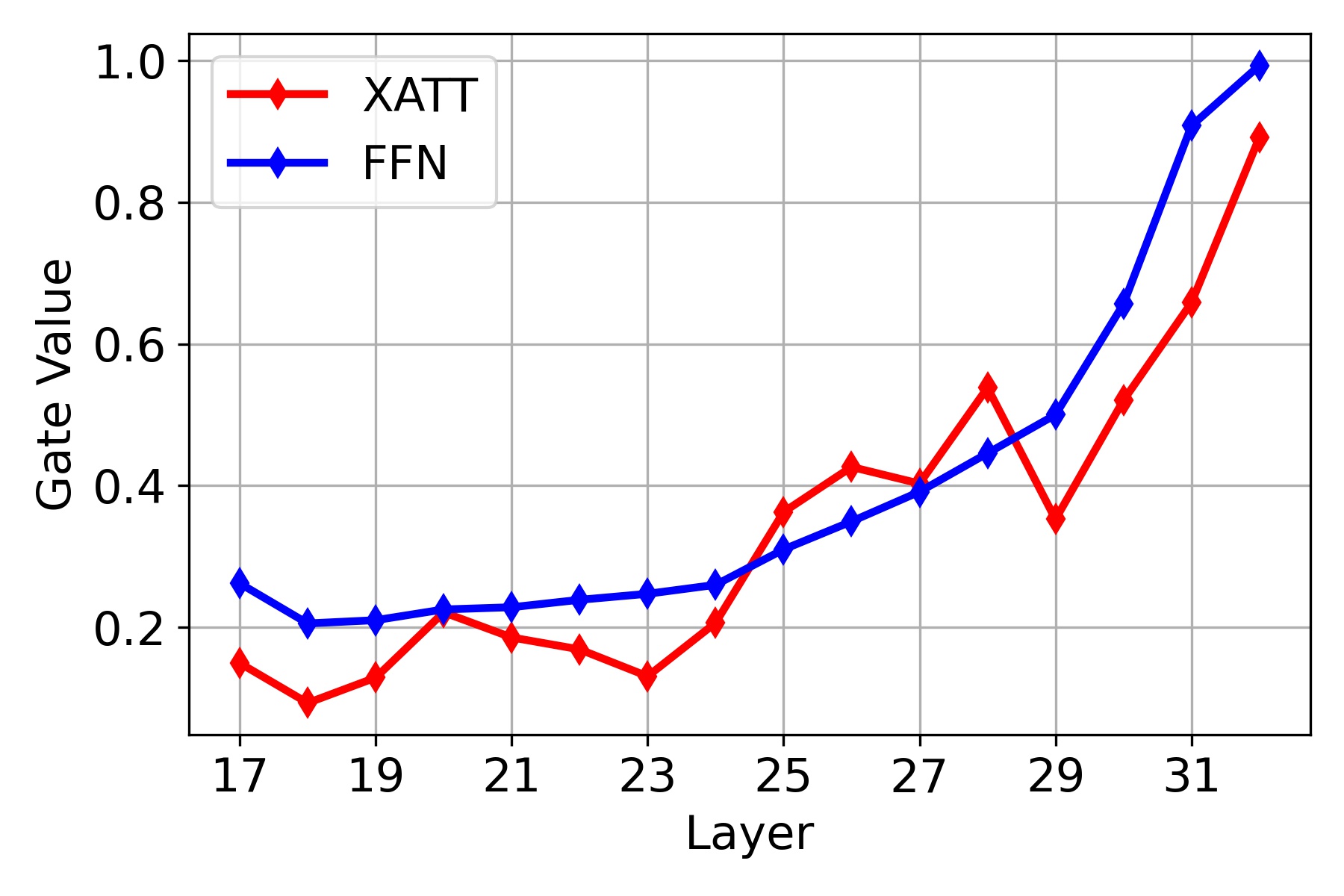}
  \caption{The absolute gate values of Gated-XATT-FFN module at different layers.}
  \label{fig:gate}
\end{figure}

We evaluated deep LLM-fusion method with the LLaMA model frozen or the last LLaMA fine-tuned. The fine-tuning was motivated by the finding that deeper layers play a more crucial role in ASR, as indicated by the increase in absolute gate values with layer depth (Figure~\ref{fig:gate}). Because of the adaptation of the last LLaMA's layer to ASR task and more trainable parameters, WERs were consistently improved under all circumstances in Table~\ref{table:deep2}.

\begin{table}[t]
\centering
\begin{tabular}{c | c | c  c} 
 \hline
 & & \multicolumn{2}{c}{\textbf{Recall}}\\
\textbf{Prompt} & \textbf{WER} & \textbf{Entity} & \textbf{OOV}\\
\hline
\multicolumn{4}{c}{\textit{SPGISpeech Dataset}} \\ 
\hline
\texttimes & 12.00 / 11.79 & 0.53 / 0.55 & 0.55 / 0.56 \\
description & \textbf{11.15} / \textbf{10.88} & \textbf{0.55} / \textbf{0.59} & \textbf{0.58} / \textbf{0.59} \\
\hline
\multicolumn{4}{c}{\textit{TedLium-2 Dataset}} \\ 
\hline
\texttimes & 9.16 / 8.99 & 0.55 / 0.57 & 0.50 / 0.50 \\
title & \textbf{8.95} / 8.80 & 0.57 / 0.59 & 0.54 / 0.53 \\
description & 9.07 / \textbf{8.76} & \textbf{0.61} / \textbf{0.65} & \textbf{0.55} / \textbf{0.56} \\
history-gt & 8.74 / \textbf{8.54} & \textbf{0.62} / 0.61 & \textbf{0.55} / \textbf{0.56} \\
history-hyp & \textbf{8.71} / 8.60 & 0.60 / 0.60 & 0.54 / 0.55 \\
 \hline
\end{tabular}
\caption{Domain adaptation performance of deep LLM-fusion. Results were provided for the model with \textbf{the last LLaMA layer frozen / fine-tuned}, respectively.}
\label{table:deep2}
\end{table}

Deep LLM-fusion outperforms the HuBERT CTC baselines without prompts, as shown in Table~\ref{table:deep2}. Prompting has a similar positive effect on deep LLM-fusion as second-pass reranking. On SPGISpeech datasets, prompting resulted in absolute WER reductions of 0.85\% and 0.91\% when the last LLaMA layer was frozen or fine-tuned, respectively. The improvements on the TedLium-2 dataset were less significant in terms of WER. When the last LLaMA layer was not fine-tuned, using a short prompt of the video title slightly outperformed a longer prompt of the video description. However, after fine-tuning the last LLaMA layer, the performance with a longer prompt improved more significantly. Additionally, by using history transcription, we achieved lower WERs compared with using global descriptive prompts. This is because previous transcriptions are more closely related to the current utterance and align with the prompt type used in long-form training. More importantly, deep LLM-fusion contributed to significantly better entity and OOV recalls. On TedLium-2, the highest entity and OOV recalls of deep LLM-fusion are 0.65 and 0.56 respectively, however, for second-pass reranking, the best recalls are only 0.51 and 0.45. This indicates that deep LLM-fusion is better at recognizing important topic-related words.
\section{Conclusion}

In our paper, we explored zero-shot domain adaptation of ASR models using two frameworks: second-pass reranking and deep LLM-fusion. By leveraging large-scale LLaMA, we achieved effective adaptation through prompt-based methods, eliminating the need for target domain data for fine-tuning. The power of LLM makes the design of prompts simple and flexible, allowing for the use of topic words or longer descriptions as prompts. The second-pass reranking method has the advantage of adapting existing ASR models without re-training. Deep LLM-fusion requires joint training of the speech encoder and LLM in the source domain. Though it requires more computational resources, it is able to recover more entity and OOV words in a new domain. In future works, we plan to perform efficient fine-tuning of the whole LLaMA model and use larger versions of LLaMA.

%
\clearpage
\bibliographystyle{IEEEbib}
\bibliography{refs}

\end{document}